\documentclass[letterpaper]{article} 
\usepackage{aaai2027}  
\usepackage[hyphens]{url}  
\usepackage{multirow}
\usepackage{graphicx} 
\usepackage{subcaption}
\usepackage{natbib}  
\usepackage{caption} 
\usepackage{algorithm}
\usepackage{algorithmic}
\usepackage{amssymb}
\usepackage{amsmath}
\usepackage{amsfonts}
\usepackage{booktabs}      
\usepackage{multirow}      
\usepackage{graphicx}      
\usepackage{array}         
\usepackage{caption}       
\usepackage{pifont}
\usepackage{newfloat}
\usepackage{listings}
\DeclareCaptionStyle{ruled}{labelfont=normalfont,labelsep=colon,strut=off} 
\floatstyle{ruled}
\newfloat{listing}{tb}{lst}{}
\floatname{listing}{Listing}

\usepackage{booktabs}

\title{DPC-Net: Dual-Prior Collaborative Network for All-in-One Image Restoration}
\author{
    Zhaokun He\textsuperscript{\rm 1}, 
    Kangbiao Shi\textsuperscript{\rm 1},
    Axi Niu\textsuperscript{\rm 1}, 
    Jian Jin\textsuperscript{\rm 2},
    Peng Wu\textsuperscript{\rm 1},
    Wei Dong\textsuperscript{\rm 3},
    Qingsen Yan\textsuperscript{\rm 1}\thanks{Corresponding author.}
}
\affiliations{
\textsuperscript{\rm 1}Northwestern Polytechnical University \\
\textsuperscript{\rm 2}Singapore Management University \\
\textsuperscript{\rm 3}Xi’an University of Architecture and Technology \\
}
\begin{document}
\maketitle
\begin{abstract}
All-in-One Image Restoration (AiOIR) aims to handle diverse degradations within a unified model. However, existing methods often overlook image semantics in degradation modeling and lack low-level visual priors during reconstruction, leading to structural distortions and semantic inconsistencies. To address these issues, we propose a novel Dual-Prior Collaborative Network (DPC-Net), which achieves high-quality restoration by jointly exploiting degradation-semantic coupled priors and low-level visual priors. Specifically, degraded images are fed into a Degradation-Aware Network (DAN) to extract degradation-semantic coupled features. To this end, a Vision-Language Model (VLM) supervises DAN by constraining its features distribution, introducing image semantics into the encoding of degradation patterns. A Degradation-Semantic Modulation Module (DSMM) further translates this guidance into degradation-semantic coupling and propagates coupled representations to the decoder. During decoding, knowledge bases provide low-level visual priors, and the Dual-Prior Collaborative Reconstruction Module (DPCR) integrates dual-prior information to guide degradation removal while preserving structure and semantics, producing high-fidelity restored images. Extensive experiments on multiple restoration benchmarks demonstrate that DPC-Net achieves superior performance against state-of-the-art AiOIR methods.
\end{abstract}

\section{Introduction}
\label{Introduction}
Image restoration, as a fundamental task in computer vision, has been extensively studied. Early efforts primarily focused on addressing a single type of degradation, such as noise~\cite{huang2021neighbor2neighbor, lin2023unsupervised}, haze~\cite{wu2021contrastive, qin2020ffa}, rain~\cite{yang2020single, chen2021robust}, or blur~\cite{cho2021rethinking, zhang2020deblurring}. Although these methods achieve promising performance under individual degradations, they struggle to generalize to complex multi-degradation scenarios. Consequently, recent research has shifted toward multi-degradation image restoration models~\cite{mou2022deep, zamir2022restormer, guo2024mambair}, which have achieved state-of-the-art performance for known combinations of degradations. However, such methods typically require a separate network for each degradation type, leading to large model sizes and substantial computational overhead.

\begin{figure}[!t]
\centering
\includegraphics[width=0.37\textwidth]{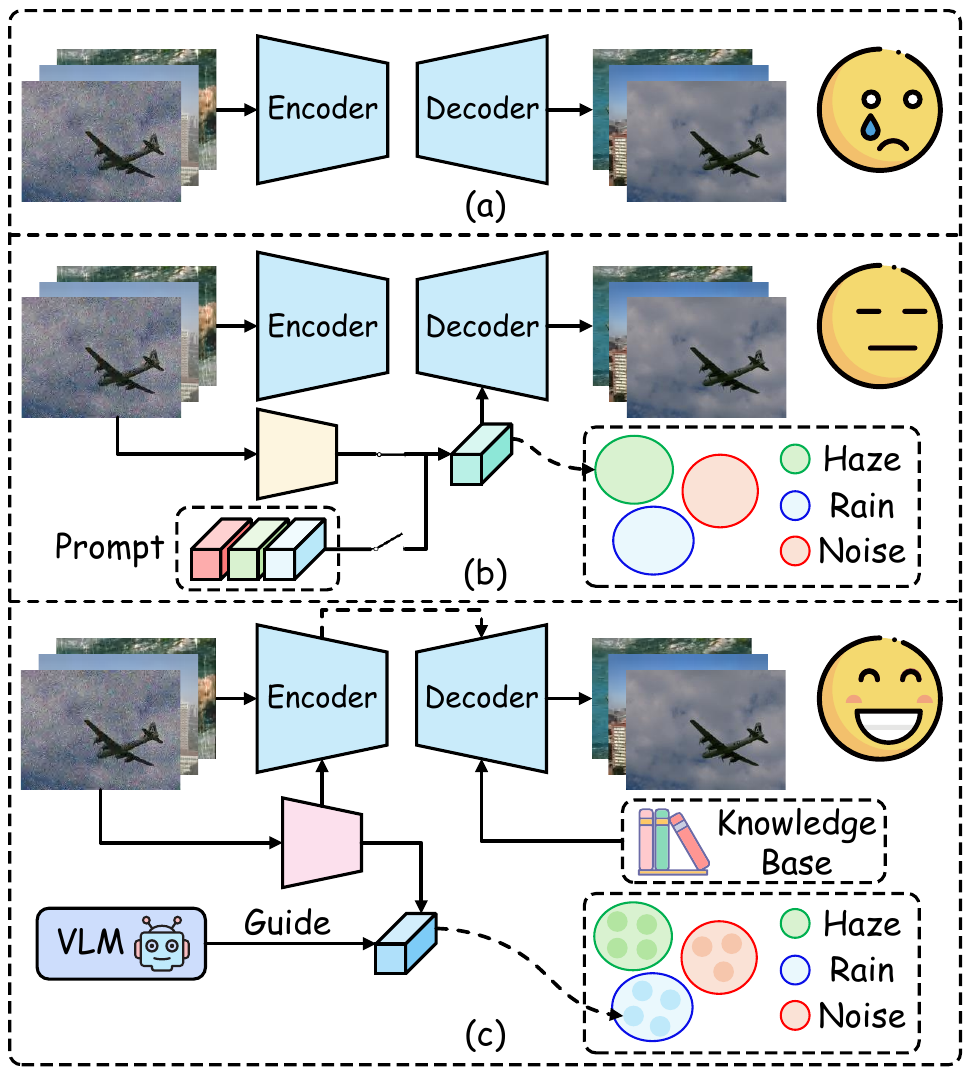}
\caption{\footnotesize Motivations of our method. (a) Models that implicitly learn degradation features. (b) Models that explicitly learn degradation features. (c) Our method.}
\label{Fig: Motivation}
\vspace{-4mm}
\end{figure}

Recently, all-in-one approaches~\cite{tang2026diffusion, zhang2026clearair} have attracted increasing attention by addressing multiple image degradations within a unified model. In general, existing all-in-one models can be broadly categorized into two types. As shown in Fig.~\ref{Fig: Motivation} (a), the first type implicitly learns degradation features~\cite{liu2022tape, ye2023adverse, cui2025adair}, relying on the network itself to automatically infer degradation patterns from degraded images without requiring explicit degradation priors. However, when confronted with diverse degradation patterns, such models often fail to accurately infer the underlying degradation type, resulting in suboptimal restoration performance. As shown in Fig.~\ref{Fig: Motivation} (b), the second type explicitly learns degradation features~\cite{zhang2025perceive, tian2025degradation, wang2026retrieve}. These methods typically introduce a lightweight auxiliary network or leverage prompts to improve the controllability of the restoration process. Although they can recognize the type of degradation, they fail to effectively couple image semantics with degradation modeling. Consequently, the model cannot understand how degradations visually distort image content. Moreover, during image reconstruction, existing methods tend to focus on the inverse removal of degradations while neglecting low-level visual priors, such as brightness, color, and edges, which lead to structural distortions or semantic inconsistencies in the restored images.

To address the aforementioned issues, we propose a novel Dual-Prior Collaborative Network (DPC-Net). Initially, we feed degraded images into the Degradation-Aware Network (DAN) to extract degradation-semantic coupled features. Specifically, a vision-language model (VLM) supervises DAN by constraining its feature distribution, thereby introducing scene semantics into the encoding of diverse degradation patterns, such as blur, noise, and low-light conditions. The Degradation-Semantic Modulation Module (DSMM) then leverages this guidance to couple degradation and semantic information and propagates the resulting coupled representations to the decoder, thereby guiding the image restoration process. In the decoding stage, we construct multiple knowledge bases with different prior information and retrieve low-level visual priors through queries. The retrieved priors, together with the degradation-semantic coupled priors output by the DSMM, are then injected into the Dual-Prior Collaborative Reconstruction Module (DPCR). This module can collaboratively leverage dual-prior information, enabling the image reconstruction process to fully incorporate priors such as luminance, color, and edges while removing degradations, thereby yielding restored results with more reasonable structures and better semantic consistency.

The main contributions of this work are summarized as follows:
\begin{itemize}
\item We propose an innovative Dual-Prior Collaborative Restoration Network, which achieves high-quality image restoration by jointly exploiting degradation-semantic coupled features and low-level visual priors.
\item A VLM-supervised Degradation-Aware Network, together with the Degradation-Semantic Modulation Module, is introduced to learn degradation-semantic coupled features, thereby guiding the image restoration process.
\item Multiple knowledge bases storing low-level visual priors are constructed to provide prior information, while employing the Dual-Prior Collaborative Reconstruction Module for dual-prior collaborative reconstruction to achieve structurally consistent restoration.
\item Extensive experiments show that our network achieves superior performance across various restoration tasks, effectively removing degradation and restoring high-quality images with coherent structures and semantics.
\end{itemize}


\section{Related Work}
\label{Related Work}
\subsection{All-in-One Image Restoration}

\textbf{Implicit Learning of Degraded Features.}
Models that implicitly learn degradation features \cite{chen2022learning, liu2022tape, ye2023adverse, li2021efficient, cui2025adair} rely on the representation capability of the network itself to adaptively perceive degradation types and distributions from degraded images, without explicitly constructing degradation priors. In recent years, some methods have improved model adaptability to multiple degradations through multi-task pretraining \cite{liu2022tape}, knowledge distillation \cite{chen2022learning}, or adaptive feature modulation \cite{cui2025adair}. For example, TAPE \cite{liu2022tape} learns general priors through task-agnostic pretraining. 
AdaIR \cite{cui2025adair} exploits frequency-domain differences to realize unified image restoration. However, such methods still depend on the network’s implicit inference of degradation features and lack explicit modeling, which leads to limitations such as insufficient representation capacity and weak interpretability in complex or unknown scenarios.

\begin{figure*}[!htp]
\centering
\includegraphics[width=0.95\textwidth]{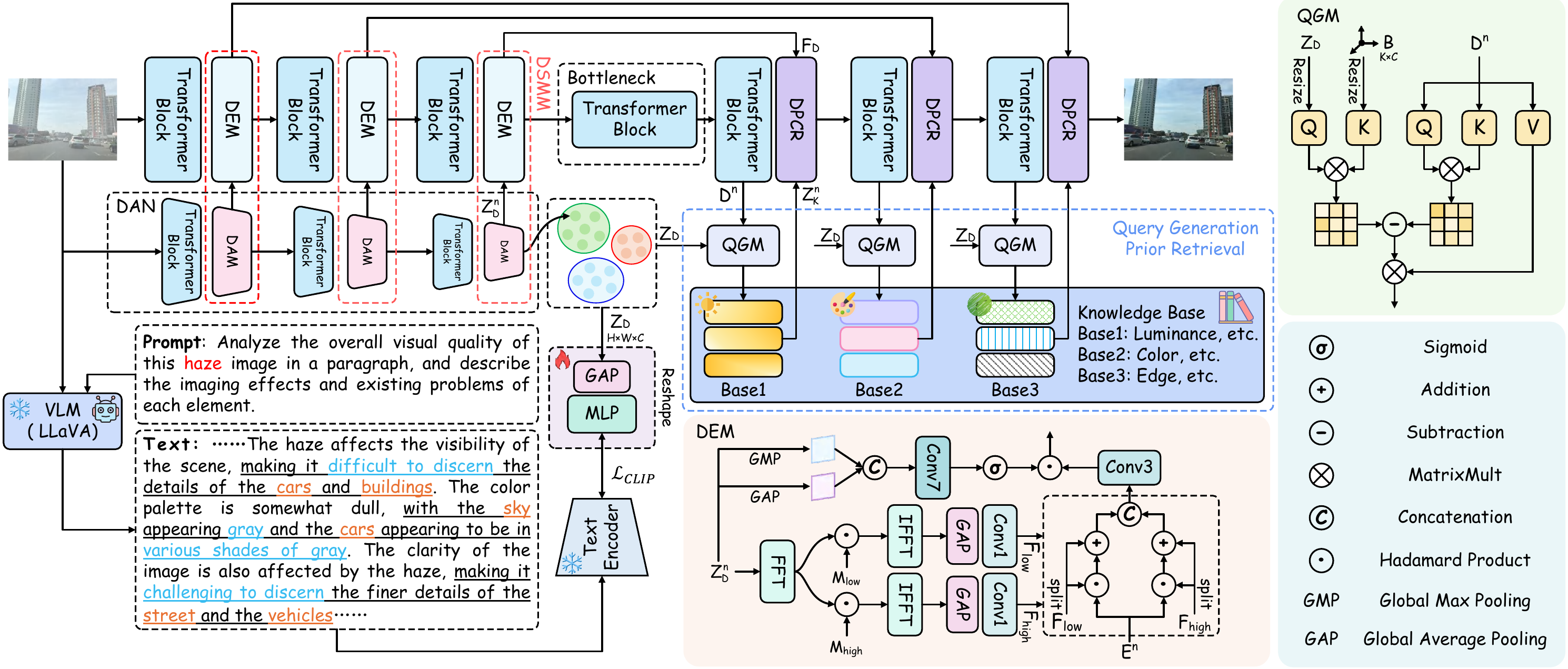}
\caption{\footnotesize Overview of our proposed Dual-Prior Collaborative Network (DPC-Net) for all-in-one image restoration.}
\label{Fig: Network}
\end{figure*}

\textbf{Explicitly Learning Degraded Features.}
Explicit degradation modeling approaches \cite{potlapalli2023promptir, yao2024neural, zhang2025perceive, tian2025degradation, zhang2026clearair, wang2026retrieve} aim to enhance model adaptability across diverse degradation scenarios by incorporating additional degradation-aware networks or learnable prompts, which explicitly encode degradation information into the image restoration pipeline. Typically, these methods first identify and characterize degradations before injecting them into the recovery process. For instance, 
Perceive-IR \cite{zhang2025perceive} jointly infers degradation types and severity levels through quality-aware and semantics-guided learning. DFPIR \cite{tian2025degradation} adapts unified parameter spaces via degradation-prompted feature perturbations. Retrieve-to-Restore \cite{wang2026retrieve} leverages a degradation prior library for retrieval-based restoration. 
However, such methods still largely confine degradation modeling to the type level, failing to integrate high-level semantic context. Moreover, while prioritizing the inverse removal of degradation components, they overlook the exploitation of crucial content priors, such as luminance, color, and edge structures, that could further inform the restoration process.

\subsection{Vision-Language Models for Image Restoration}
Vision-Language Models (VLMs) introduce cross-modal semantic priors, providing additional content-aware and degradation-aware guidance for image restoration. In recent years, several methods \cite{zhou2025low, qu2024xpsr, sun2026adapting, cheng2026unildiff, dong2026learning} have begun to explore the use of VLMs in image restoration tasks. For example, 
DATPRL-IR \cite{dong2026learning} uses a VLM to produce multi-dimensional descriptions of image content, color, and brightness. UniLDiff \cite{cheng2026unildiff} adopts a VLM to describe image degradation types and content. In addition, XPSR \cite{qu2024xpsr} extracts high-level semantic descriptions with a VLM to characterize image content and spatial layout, while also extracting low-level semantic descriptions to depict degradations and quality defects. Although the low-level semantic descriptions integrate image semantics with degradation information, they primarily capture coarse-grained semantics and degradation patterns of salient objects in the scene, lacking the fine-grained details required for pixel-level reconstruction. As a result, they can hardly provide effective guidance for image restoration through direct injection into the decoder. Consequently, we inject hierarchical features from the DAN into the DEM at each encoder level, thereby obtaining superior priors.

\section{Method}
\label{Method}
\subsection{Overall Pipeline}
To enable the model to better capture the visual distortions caused by degradation during degradation modeling while more effectively leveraging low-level visual priors for image reconstruction, we propose the Dual-Prior Collaborative Network (DPC-Net). As shown in Fig.~\ref{Fig: Network}, we first feed degraded images into the Degradation-Aware Network (DAN) to extract degradation-semantic coupled features. To this end, a Vision-Language Model (VLM) supervises DAN by constraining its feature distribution, thereby introducing scene semantics into the encoding of diverse degradation patterns, such as blur, noise, and low-light conditions. Guided by the VLM, the Degradation-Semantic Modulation Module (DSMM) couples degradation and semantic information. Specifically, hierarchical degradation features from DAN are injected into the corresponding encoder levels via the Degradation Embedding Modulation Module (DEM), and the modulated features are transmitted to the decoder. During the decoding stage, we construct multiple knowledge bases containing diverse low-level visual priors. By querying these bases, we supply the reconstruction process with rich low-level visual priors. Finally, the dual-prior information is jointly injected into the Dual-Prior Collaborative Reconstruction Module (DPCR), where their synergy guides the image restoration process, thereby removing complex degradations while generating high-quality restored images with more plausible structures and better semantic consistency. The details of each component in the proposed network are presented in the following subsections.

\subsection{Integrating VLMs into Image Restoration}
Existing methods fail to consider image semantics during degradation modeling, making it difficult for the model to understand how degradations visually distort image content. Meanwhile, VLMs can provide cross-modal semantic descriptions of degraded images, capturing scene content, imaging quality, color variations, and detail degradation. Compared with manually defined degradation labels, such textual descriptions offer a more expressive way to characterize how degradations affect visual appearance. Therefore, we leverage VLMs to guide the image restoration process. The overall process will be described in detail below.

\textbf{Degradation-Aware Network.}
Given a degraded image $\mathbf{I}_d$, we construct an image quality analysis prompt to guide the VLM to generate textual descriptions from multiple perspectives, including global visual quality, local imaging artifacts, scene content, color representation, and detail degradation. Subsequently, $\mathbf{I}_{d}$ is fed into the DAN for feature extraction. Specifically, the Distribution-aware Normalization within the Degradation-Aware Module (DAM) adaptively adjusts the feature distributions based on different degradation types, thereby accommodating the variations in feature statistics caused by diverse degradations. On this basis, the degradation-semantic coupled features produced by DAN are further constrained by the textual descriptions generated by the VLM, enabling them to encode degradation patterns while incorporating scene semantic information. The overall process can be formulated as follows:
\begin{equation}
\mathbf{Z}_{D}^{n} = \mathrm{SA}(\mathrm{DNorm}(\mathbf{Z}_{D}^{n})) + \mathbf{Z}_{D}^{n},
\end{equation}
\begin{equation}
\mathbf{Z}_{D}^{n+1} = \mathrm{FFN}(\mathrm{DNorm}(\mathbf{Z}_{D}^{n})) + \mathbf{Z}_{D}^{n}
\end{equation}
Where $\mathrm{DNorm}(\cdot)$ denotes Distribution-aware Normalization, $\mathrm{SA}(\cdot)$ denotes self-attention, $\mathrm{FFN}(\cdot)$ denotes a feedforward network, and $\mathbf{Z}_{D}^{n}$ denotes the feature of the $n$-th layer of the DAN. Details of DAN are provided in the \textbf{Appendix}.

\begin{figure}[t]
\centering
\includegraphics[width=0.45\textwidth]{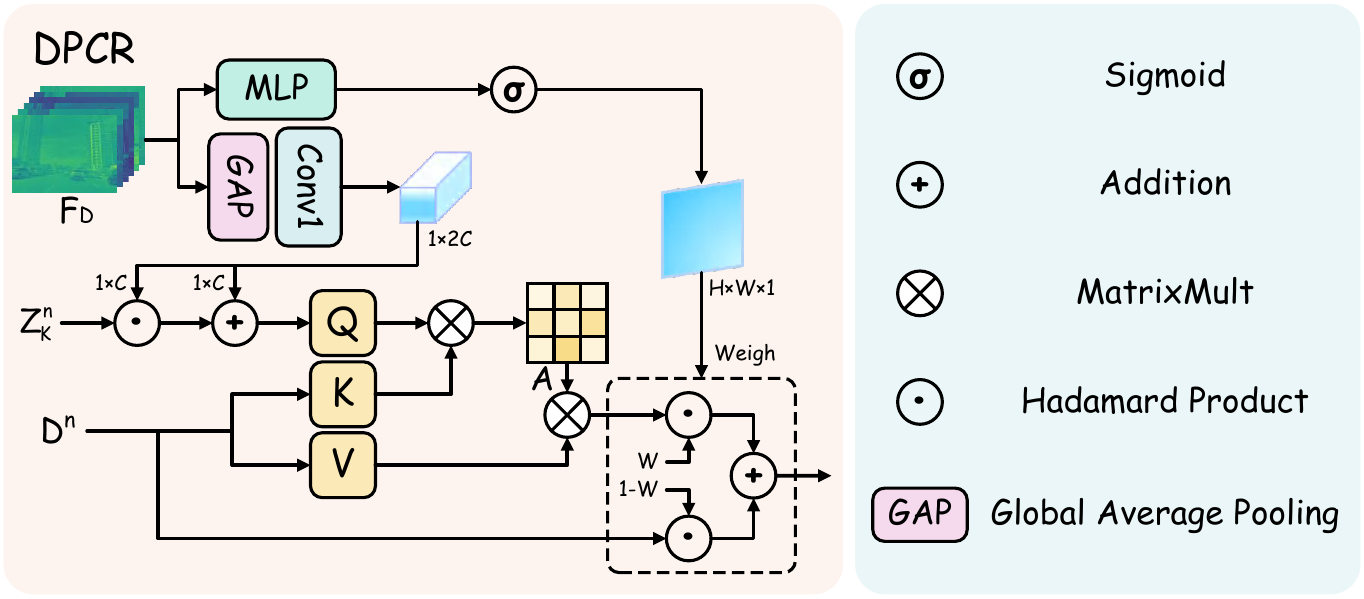}
\caption{\footnotesize Illustration of our proposed Dual-Prior Collaborative Reconstruction Module (DPCR).}
\label{Fig: Module}
\end{figure}

\begin{figure}[t]
\centering
\includegraphics[width=0.45\textwidth]{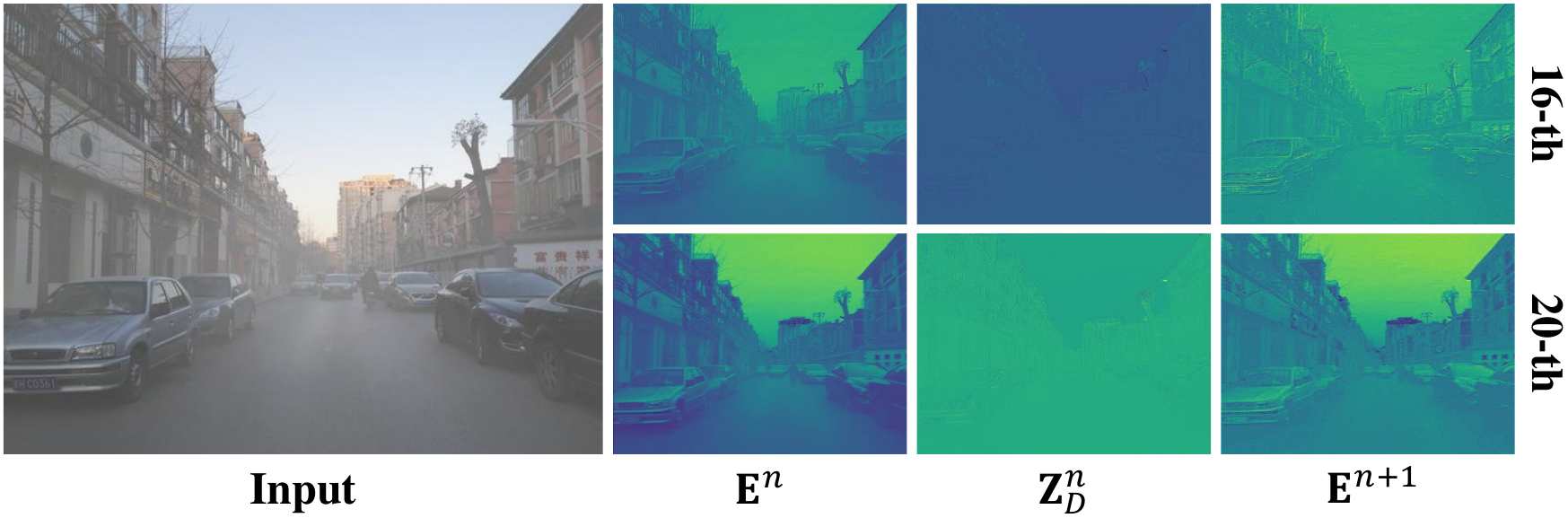}
\caption{\footnotesize Comparison of feature maps across different channels.}
\label{Fig: Feature Map}
\end{figure}

\textbf{Degradation Embedding Modulation Module.}
As shown in Fig.~\ref{Fig: Feature Map}, although the degradation-semantic coupled features extracted by DAN can capture coarse-grained subject semantics and overall degradation patterns, they lack fine-grained spatial information and thus cannot be directly injected into the decoder to guide pixel-level reconstruction. To address this issue, we inject the hierarchical feature $\mathbf{Z}_{D}^{n}$ from the DAN into the corresponding level of the DEM within the encoder. Considering that different degradations exhibit distinct characteristic distributions in the frequency domain \cite{cui2025adair}, we first decompose $\mathbf{Z}_{D}^{n}$ into high- and low-frequency features within the DEM:
\begin{equation}
\mathbf{F}_{d} = \mathrm{Conv1}(\mathrm{GAP}(\mathrm{IFFT}(\mathrm{FFT}(\mathbf{Z}_{D}^{n}) \odot \mathbf{M}_{d})))
\end{equation}
where $d\in\{low, high\}$ indicates the frequency domain. Subsequently, the high- and low-frequency features are interacted with the encoder hierarchical feature $\mathbf{E}^{n}$ separately, followed by their fusion:
\begin{equation}
\mathbf{E}^{n}_{d} = \alpha_{d} \odot \mathbf{E}^{n} + \beta_{d}
\end{equation}
\begin{equation}
\mathbf{E}^{n}_{fused} = \mathrm{Conv3}([\mathbf{E}^{n}_{high}:\mathbf{E}^{n}_{low}])
\end{equation}
where $\alpha_{d}$ and $\beta_{d}$ denote the results of splitting $\mathbf{F}_{d}$ along the channel dimension, and $[:]$ indicates channel-wise concatenation. Finally, to further establish the correspondence between semantic regions and degradation, we generate a spatial attention map $\mathbf{A}_{s}$ from $\mathbf{Z}_{D}^{n}$ to adaptively weight $\mathbf{E}^{n}_{fused}$:
\begin{equation}
\mathbf{A}_{s} = \sigma(\mathrm{Conv7}([\mathrm{GMP}(\mathbf{Z}_{D}^{n}):\mathrm{GAP}(\mathbf{Z}_{D}^{n})]))
\end{equation}
\begin{equation}
\mathbf{E}^{n+1} = \mathbf{A}_{s} \odot \mathbf{E}^{n}_{fused}
\end{equation}

\subsection{Leveraging Knowledge Bases for Image Restoration}
Degradation-semantic coupling features primarily characterize the visual distortion effects of degradation on image content. However, high-quality reconstruction requires not only modeling the correlation between degradation and semantic content but also incorporating low-level visual priors to guide reconstruction. For different restoration tasks, the roles of low-level visual priors vary. In low-light enhancement, brightness and color priors help restore reasonable exposure and color naturalness. In dehazing, color and edge priors facilitate the recovery of natural colors and long-range structures. In denoising, deraining, and deblurring, edge priors help preserve genuine structural boundaries. Based on this, we set up a learnable knowledge base at each layer of the decoder. Since features at different layers possess varying spatial scales and semantic granularities, the corresponding knowledge bases can adaptively learn multi-scale low-level visual priors. Furthermore, we equip each knowledge base with a Query Generation Module (QGM) to generate queries. Its detailed design is elaborated below.

\textbf{Query Generation Module.}
Since the decoder's hierarchical features contain degradation information, directly using them as queries introduces interference. Therefore, we leverage the degradation-semantic coupled features to eliminate this interference and generate clear queries. Inspired by \cite{ye2025differential}, we introduce a differential attention mechanism to eliminate interference by leveraging the degradation-semantic coupled features. However, these degradation-semantic coupled features still retain image-semantic information, which may impair the semantic integrity of the query features. To resolve this issue, we construct a set of degradation basis vectors $\mathbf{B} \in \mathbb{R}^{K \times C}$ and impose a loss function to enforce pairwise orthogonality among them. Subsequently, we leverage the degradation-semantic coupled features to query $\mathbf{B}$, thereby filtering out the embedded semantic information:
\begin{equation}
\mathbf{Q}_{deg} = \mathrm{Conv3}(\mathrm{Interp}(\mathbf{Z}_{D}, (H,W)))
\end{equation}
\begin{equation}
\mathbf{K}_{base} = \mathrm{Conv3}(\mathrm{Reshape}(\mathrm{Interp}(\mathbf{B}, H \cdot W)))
\end{equation}
\begin{equation}
\mathbf{A}_{deg} = \mathbf{Q}_{deg} \otimes \mathbf{K}_{base}
\end{equation}
where $\mathrm{Interp}(\cdot, \cdot)$ represents the operation of interpolating features to a specified shape. Finally, we integrate the differential attention mechanism to effectively eliminate the interfering components in the hierarchical decoder features, thereby generating high-quality clean queries:
\begin{equation}
\mathbf{A}_{diff} = \mathbf{A}_{feat} - \mathbf{A}_{deg}
\end{equation}
\begin{equation}
\mathbf{F}_{Q} = \mathrm{Softmax}(\mathbf{A}_{diff}) \otimes \mathbf{V}_{feat}
\end{equation}
Where $\mathbf{A}_{feat}$ and $\mathbf{V}_{feat}$ are derived by applying a self-attention mechanism to the decoder-level features $\mathbf{D}^n$, and $\mathbf{F}_{Q}$ denotes the generated clean query.

\begin{table*}[t]
\centering
\small
\setlength\tabcolsep{1.5pt} 
\begin{tabular}{lcccccccccccccc}
\toprule[1pt]
\multirow{2}{*}{Method} & \multirow{2}{*}{Source} & \multirow{2}{*}{Params.} 
    & \multicolumn{2}{c}{\textit{Dehazing}}
    & \multicolumn{2}{c}{\textit{Deraining}}
    & \multicolumn{6}{c}{\textit{Denoising}}
    & \multicolumn{2}{c}{\multirow{2}{*}{Average}}  \\
    \cmidrule(lr){4-5} \cmidrule(lr){6-7} \cmidrule(lr){8-13} 
    &
    &
    & \multicolumn{2}{c}{SOTS} 
    & \multicolumn{2}{c}{Rain100L} 
    & \multicolumn{2}{c}{BSD68\textsubscript{$\sigma$=15}} 
    & \multicolumn{2}{c}{BSD68\textsubscript{$\sigma$=25}} 
    & \multicolumn{2}{c}{BSD68\textsubscript{$\sigma$=50}} 
    &  \\
\midrule
    AirNet \cite{li2022all} & CVPR'22 & 9M
        & 27.94 & .962
        & 34.90 & .967
        & 33.92 & .933
        & 31.26 & .888
        & 28.00 & .797
        & 31.20 & .910 \\
    IDR \cite{zhang2023ingredient} & CVPR'23 & 15M
        & 29.87 & .970
        & 36.03 & .971
        & 33.89 & .931
        & 31.32 & .884
        & 28.04 & .798
        & 31.83 & .911 \\
    PromptIR \cite{potlapalli2023promptir} & NeurIPS'23 & 33M
        & 30.58 & .974
        & 36.37 & .972
        & 33.98 & .933
        & 31.31 & .888
        & 28.06 & .799
        & 32.06 & .913 \\
    NDR \cite{yao2024neural} & TIP'24 & 28M
        & 28.64 & .962 
        & 35.42 & .969 
        & 34.01 & .932 
        & 31.36 & .887 
        & 28.10 & .798 
        & 31.51 & .910 \\
    Gridformer \cite{wang2024gridformer} & IJCV'24 & 34M 
        & 30.37 & .970 
        & 37.15 & .972 
        & 33.93 & .931 
        & 31.37 & .887 
        & 28.11 & .801 
        & 32.19 & .912 \\
    InstructIR \cite{conde2024instructir} & ECCV'24 & 16M 
        & 30.22 & .959
        & 37.98 & .978
        & 34.15 & .933
        & 31.52 & .890 
        & 28.30 & .803
        & 32.43 & .913 \\
    Up-Restorer \cite{liu2025up} & AAAI'25 & 28M
        & 30.68 & .977
        & 36.74 & .978
        & 33.99 & .933
        & 31.33 & .888
        & 28.07 & .799
        & 32.16 & .915 \\
    Perceive-IR \cite{zhang2025perceive} & TIP'25 & 42M 
        & 30.87 & .975
        & 38.29 & .980
        & 34.13 & .934
        & \textcolor{blue}{31.53} & .890
        & \textcolor{blue}{28.31} & .804
        & 32.63 & .917 \\
    AdaIR \cite{cui2025adair} & ICLR'25 & 29M 
        & 31.06 & .980
        & 38.64 & .983
        & 34.12 & .935
        & 31.45 & .892
        & 28.19 & .802 
        & 32.69 & .918 \\
    R2R \cite{wang2026retrieve} & CVPR'26 & 20M
        & 31.40 & .977 
        & 37.46 & .980 
        & 34.10 & .936
        & 31.45 & \textcolor{blue}{.895} 
        & 28.22 & .806 
        & 32.53 & .918 \\
    DFPIR \cite{tian2025degradation} & CVPR'25 & 30M
        & 31.87 & .980
        & \textcolor{blue}{38.65} & .982
        & 34.14 & .935
        & 31.47 & .893
        & 28.25 & .806
        & \textcolor{blue}{32.88} & .919 \\
    VLU-Net \cite{zeng2025vision} & CVPR'25 & 35M 
        & 30.71 & .980
        & \textcolor{red}{38.93} & .984 
        & 34.13 & .935
        & 31.48 & .892 
        & 28.23 & .804
        & 32.70 & .919 \\
    StarIR \cite{cui2026starir} & TPAMI'26 & 9M
        & 30.89 & .979
        & 38.50 & \textcolor{blue}{.984}
        & 34.17 & .936
        & 31.51 & .893
        & 28.26 & .806
        & 32.67 & .920 \\
    HOGformer \cite{wu2026gradient} & AAAI'26 & 17M
        & \textcolor{blue}{31.91} & .981
        & 38.50 & .983
        & 34.04 & .935
        & 31.40 & .892
        & 28.16 & .804
        & 32.80 & .919 \\
    DRNet \cite{li2026drnet} & TMM'26 & 7M 
        & 31.15 & .979 
        & 38.28 & .983 
        & \textcolor{red}{34.20} & \textcolor{blue}{.937} 
        & \textcolor{red}{31.55} & .894 
        & 28.27 & \textcolor{blue}{.807} 
        & 32.69 & \textcolor{blue}{.920}  \\
    ClearAIR \cite{zhang2026clearair} & AAAI'26 & 31M
        & 31.08 & \textcolor{blue}{.981}
        & 38.61 & \textcolor{red}{.984}
        & \textcolor{blue}{34.18} & .935 
        & 31.50 & .891 
        & \textcolor{red}{28.31} & .804 
        & 32.74 & .919 \\
\midrule
    Ours & -- & 27M
        & \textcolor{red}{32.99} & \textcolor{red}{.983}
        & 38.16 & .982
        & 34.15 & \textcolor{red}{.937}
        & 31.50 & \textcolor{red}{.896}
        & 28.26 & \textcolor{red}{.813}
        & \textcolor{red}{33.01} & \textcolor{red}{.922} \\
\bottomrule[1pt]
\end{tabular}
\caption{\footnotesize Comparison to state-of-the-art all-in-one methods on the three degradation tasks. \textcolor{red}{Best} and \textcolor{blue}{second} best performances are highlighted. PSNR (dB, $\uparrow$) and SSIM ($\uparrow$) metrics are reported on the full RGB images.}
\label{Tab: Three Degradations}
\end{table*}

\subsection{Dual-Prior Collaborative Reconstruction Module}
To leverage dual prior information for image restoration, we embed DPCR into each layer of the decoder. The feature $\mathbf{F}_{D}$ from the DEM contains rich semantic information as well as degradation cues, enabling the model to understand how the degradation distorts image content in the visual domain. Therefore, we utilize $\mathbf{F}_{D}$ to, on the one hand, select from the queried low-level visual priors the effective prior information required under the current degradation condition:
\begin{equation}
\alpha_s, \beta_s = \mathrm{chunk}(\mathrm{Conv1}(\mathrm{GAP}(\mathbf{F}_{D}))),
\end{equation}
\begin{equation}
\hat{\mathbf{Z}}_K^n = \mathbf{Z}_K^n \odot \alpha_s + \beta_s
\end{equation}
On the other hand, it adaptively measures the restoration strength of each pixel in an image:
\begin{equation}
\mathbf{W}_{s} = \sigma(\mathrm{MLP}(\mathbf{F}_{D})),
\end{equation}
\begin{equation}
\mathbf{D}^{n+1} = \mathbf{D}^n_{rec} \odot \mathbf{W}_s + \mathbf{D}^n \odot (1 - \mathbf{W}_s)
\end{equation}
Where $\mathbf{D}^n_{rec}$ is the output of restoring $\mathbf{D}^{n}$ using $\hat{\mathbf{Z}}_K^n$.

\section{Experiment}
\label{Experiment}
\subsection{Experimental Setup}
\textbf{Datasets.}
Following existing works \cite{wang2026retrieve, zhang2026clearair}, we establish two configurations involving three degradations and five degradations, respectively. For the denoising task, we merge BSD400 \cite{arbelaez2010contour} and WED \cite{ma2016waterloo} as the training set and synthesize noisy images by adding Gaussian noise with a noise level of $\sigma \in {[15,25,50]}$, while employing BSD68 \cite{martin2001database} for testing. The deraining task utilizes the Rain100L \cite{yang2017deep} dataset. The dehazing task adopts the SOTS \cite{li2018benchmarking} dataset. The deblurring task employs the GoPro \cite{nah2017deep} dataset. The low-light enhancement task uses the LOLv1 \cite{wei2018deep} dataset. Additionally, for single-task experiments, our method is trained on the respective training set. Dataset details are provided in the \textbf{Appendix}.

\textbf{Implementation Details.}
Our DPC-Net provides an end-to-end trainable solution. Leveraging LLaVA as the vision-language model and Restormer as the backbone network, the proposed architecture adopts a four-level encoder-decoder structure. Each level incorporates a distinct number of Transformer blocks, specifically configured as $[4, 6, 6, 8]$ from level-1 to level-4. Each decoder layer is equipped with a knowledge base containing $m=256$ features, whose dimensions match those of the corresponding layer. Experiments are conducted on NVIDIA GeForce RTX 4090 GPUs using PyTorch. During training, we set the learning rate to $1e^{-4}$ and the input patch size to $128^2$. Network optimization employ a composite loss function comprising $\mathcal{L}_1$, $\mathcal{L}_{edge}$, $\mathcal{L}_{SSIM}$, $\mathcal{L}_{CLIP}$, and $\mathcal{L}_{base}$ losses (detailed in the \textbf{Appendix}), in conjunction with the Adam optimizer ($\beta_1$ = 0.9, $\beta_2$ = 0.999). Training is performed on cropped images, augmented via random horizontal and vertical flips.

\begin{table*}[t]
\centering
\small
\setlength\tabcolsep{1.5pt}
\begin{tabular}{lcccccccccccccc}
\toprule[1pt]
    \multirow{2}{*}{Method} 
    & \multirow{2}{*}{Source}
    & \multirow{2}{*}{Params.} 
    & \multicolumn{2}{c}{\textit{Dehazing}} 
    & \multicolumn{2}{c}{\textit{Deraining}} 
    & \multicolumn{2}{c}{\textit{Denoising}} 
    & \multicolumn{2}{c}{\textit{Deblurring}} 
    & \multicolumn{2}{c}{\textit{Low-Light}} 
    & \multicolumn{2}{c}{\multirow{2}{*}{Average}}  
    \\
    \cmidrule(lr){4-5} 
    \cmidrule(lr){6-7} 
    \cmidrule(lr){8-9} 
    \cmidrule(lr){10-11} 
    \cmidrule(lr){12-13}
    &
    & 
    &
    \multicolumn{2}{c}{SOTS} 
    & \multicolumn{2}{c}{Rain100L} 
    & \multicolumn{2}{c}{BSD68\textsubscript{$\sigma$=25}} 
    & \multicolumn{2}{c}{GoPro} 
    & \multicolumn{2}{c}{LOLv1} 
    &  
    \\
    \midrule
    AirNet \cite{li2022all} & CVPR'22 & 9M 
    & 21.04 & .884
    & 32.98 & .951 
    & 30.91 & .882
    & 24.35 & .781
    & 18.18 & .735
    & 25.49 & .846
    \\
    IDR \cite{zhang2023ingredient} & CVPR'23 & 15M
    & 25.24 & .943
    & 35.63 & .965 
    & \textcolor{red}{31.60} & .887 
    & 27.87 & .846 
    & 21.34 & .826 
    & 28.34 & .893
    \\
    PromptIR \cite{potlapalli2023promptir} & NeurIPS'23 & 33M 
    & 26.54 & .949
    & 36.37 & .970
    & 31.47 & .886 
    & 28.71 & .881
    & 22.68 & .832
    & 29.15 & .904
    \\
    Gridformer \cite{wang2024gridformer} & IJCV'24 & 34M 
    & 26.79 & .951
    & 36.61 & .971
    & 31.45 & .885
    & 29.22 & .884
    & 22.59 & .831
    & 29.33 & .904
    \\
    InstructIR \cite{conde2024instructir} & ECCV'24 & 16M
    & 27.10 & .956
    & 36.84 & .973
    & 31.40 & .887 
    & 29.40 & .886
    & 23.00 & .836
    & 29.55 & .907
    \\
    Perceive-IR \cite{zhang2025perceive} & TIP'25 & 42M
    & 28.19 & .964
    & 37.25 & .977 
    & 31.44 & .887
    & 29.46 & .886
    & 22.81 & .833
    & 29.84 & .909 
    \\
    AdaIR \cite{cui2025adair} & ICLR'25 & 29M
    & 30.53 & .978
    & 38.02 & .981 
    & 31.35 & .889
    & 28.12 & .858
    & 23.00 & .845
    & 30.20 & .910
    \\
    VLU-Net \cite{zeng2025vision} & CVPR'25 & 35M
    & 30.84 & .980 
    & \textcolor{blue}{38.54} & .982
    & 31.43 & .891  
    & 27.46 & .840
    & 22.29 & .833
    & 30.11 & .905
    \\
    ClearAIR \cite{zhang2026clearair} & AAAI'26 & 31M 
    & 30.12 & .978 
    & 38.20 & .982
    & 31.53 & .888 
    & 29.67 & .887
    & 22.83 & .846
    & 30.45 & .916
    \\
    StarIR \cite{cui2026starir} & TPAMI’26 & 9M
    & 30.46 & .977
    & \textcolor{red}{38.69} & \textcolor{red}{.984}
    & 31.47 & .893
    & 28.63 & .871
    & 23.32 & .858
    & 30.51 & .917
    \\
    HOGformer \cite{wu2026gradient} & AAAI'26 & 17M
    & 31.16 & .979
    & 38.05 & .981
    & 31.19 & .884
    & 28.62 & .867
    & \textcolor{red}{24.46} & \textcolor{blue}{.858}
    & \textcolor{blue}{30.70} & .914
    \\
    R2R \cite{wang2026retrieve} & CVPR'26 & 20M
    & 30.64 & .974 
    & 36.61 & .975 
    & 31.35 & .891 
    & \textcolor{red}{30.93} & \textcolor{red}{.911} 
    & 22.88 & .856 
    & 30.48 & \textcolor{blue}{.921}
    \\
    DRNet \cite{li2026drnet} & TMM’26 & 7M
    & \textcolor{blue}{31.28} & \textcolor{blue}{.980}
    & 38.13 & \textcolor{blue}{.982} 
    & \textcolor{blue}{31.54} & \textcolor{blue}{.894}
    & 29.01 & .870
    & 22.30 & .846
    & 30.45 & .914
    \\
\midrule
    Ours & -- & 27M
    & \textcolor{red}{32.17} & \textcolor{red}{.980}
    & 37.90 & .981
    & 31.46 & \textcolor{red}{.894}
    & \textcolor{blue}{29.86} & \textcolor{blue}{.895}
    & \textcolor{blue}{23.60} & \textcolor{red}{.865}
    & \textcolor{red}{31.00} & \textcolor{red}{.923} \\
\bottomrule[1pt]
\end{tabular}
\caption{\footnotesize Comparison to state-of-the-art all-in-one methods on the five degradation tasks. \textcolor{red}{Best} and \textcolor{blue}{second} best performances are highlighted. PSNR (dB, $\uparrow$) and SSIM ($\uparrow$) metrics are reported on the full RGB images.}
\label{Tab: Five Degradations}
\end{table*}

\begin{figure*}[!t]
\centering
\scalebox{0.93}{
\begin{minipage}[t]{0.135\linewidth}
    \centering
    \centerline{\includegraphics[width=\textwidth]{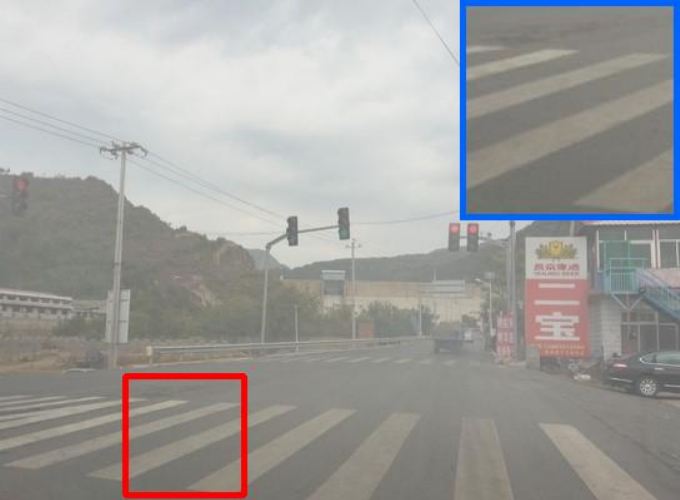}}
    \vspace{2pt}
    \centerline{\includegraphics[width=\textwidth]{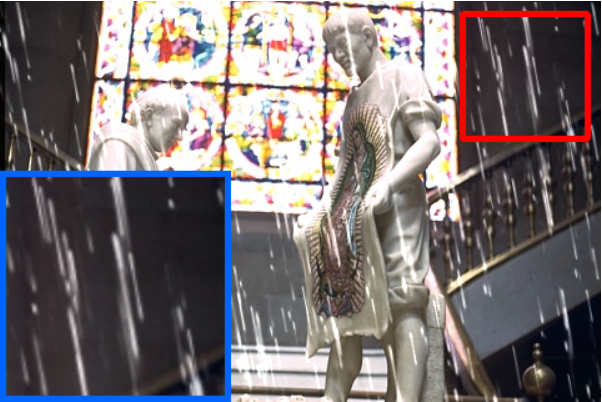}}
    \vspace{2pt}
    \centerline{\includegraphics[width=\textwidth]{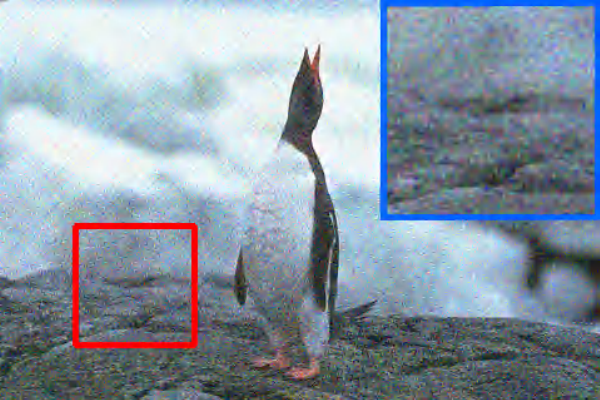}}
    \centerline{\small Input}
\end{minipage}
\begin{minipage}[t]{0.135\linewidth}
    \centering
    \centerline{\includegraphics[width=\textwidth]{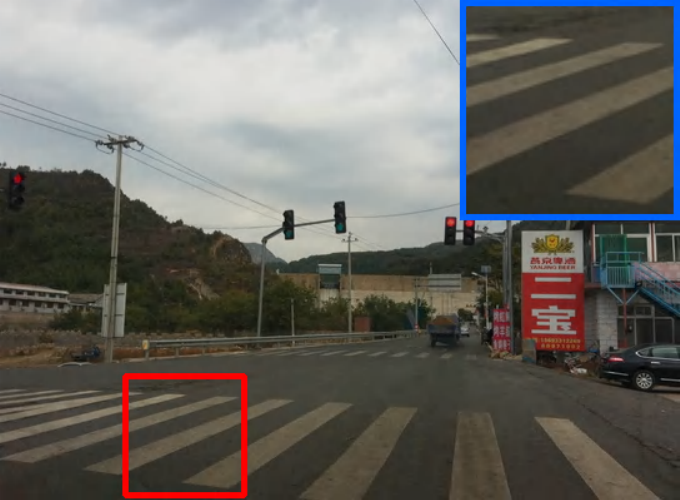}}
    \vspace{2pt}
    \centerline{\includegraphics[width=\textwidth]{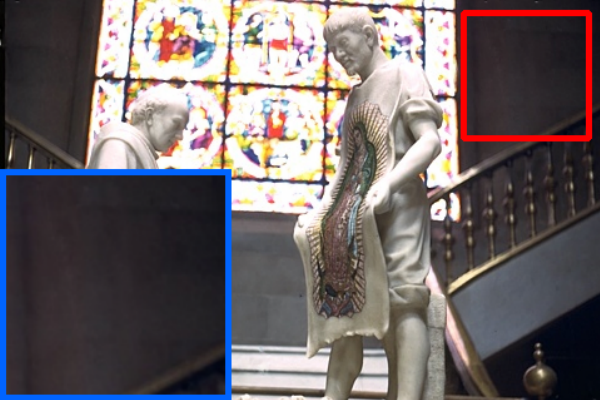}}
    \vspace{2pt}
    \centerline{\includegraphics[width=\textwidth]{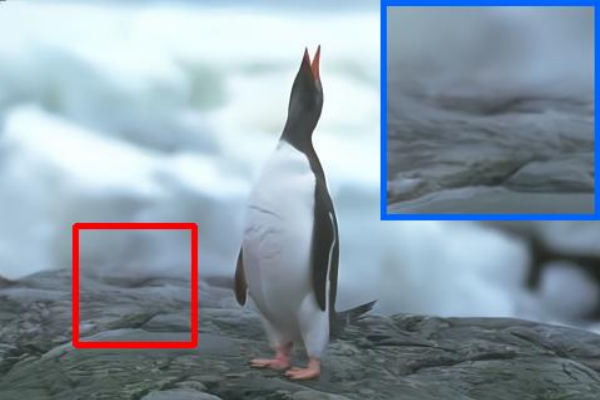}}
    \centerline{\small AdaIR}
\end{minipage}
\begin{minipage}[t]{0.135\linewidth}
    \centering
    \centerline{\includegraphics[width=\textwidth]{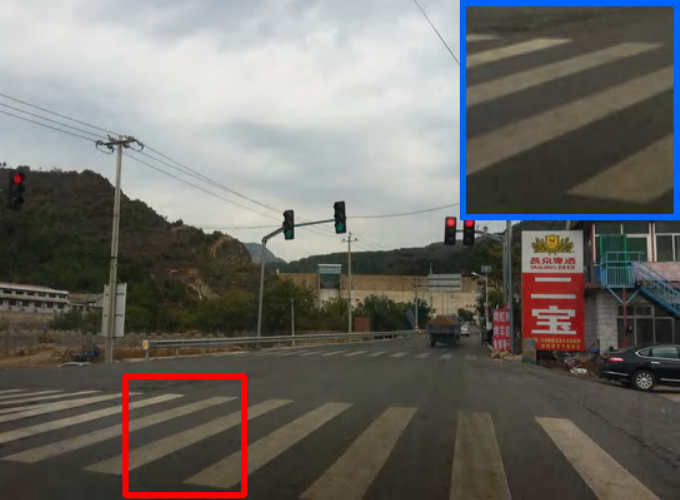}}
    \vspace{2pt}
    \centerline{\includegraphics[width=\textwidth]{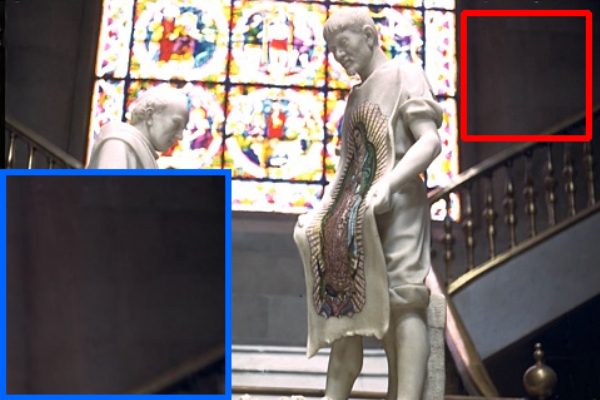}}
    \vspace{2pt}
    \centerline{\includegraphics[width=\textwidth]{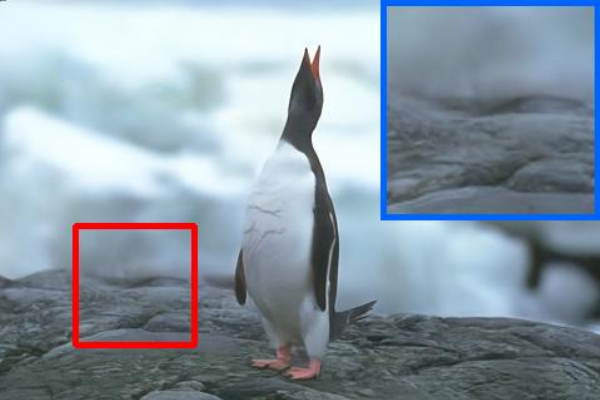}}
    \centerline{\small DFPIR}
\end{minipage}
\begin{minipage}[t]{0.135\linewidth}
    \centering
    \centerline{\includegraphics[width=\textwidth]{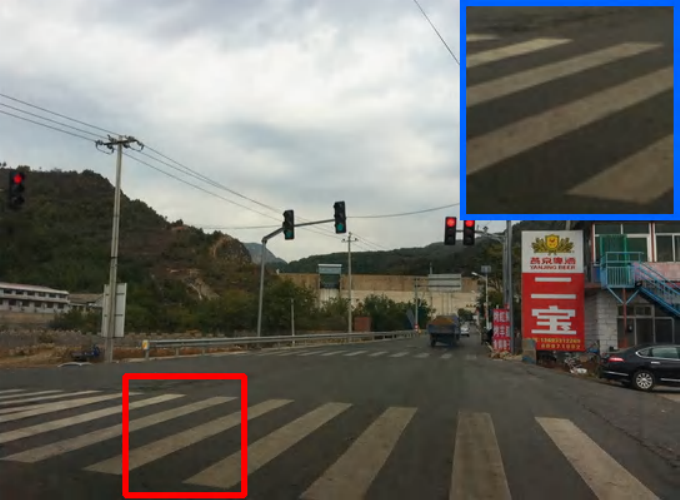}}
    \vspace{2pt}
    \centerline{\includegraphics[width=\textwidth]{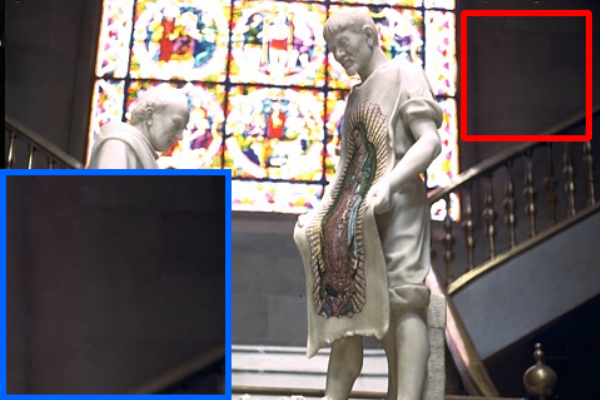}}
    \vspace{2pt}
    \centerline{\includegraphics[width=\textwidth]{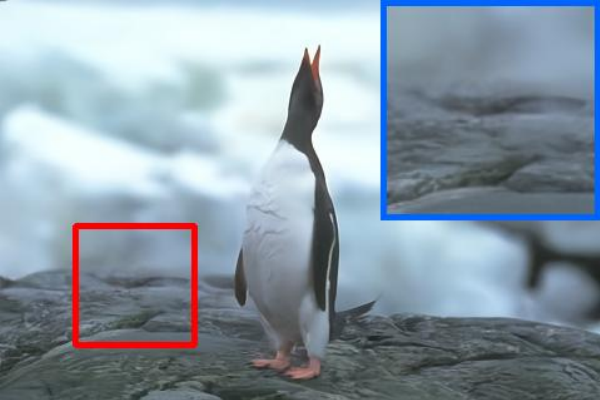}}
    \centerline{\small VLU-Net}
\end{minipage}
\begin{minipage}[t]{0.135\linewidth}
    \centering
    \centerline{\includegraphics[width=\textwidth]{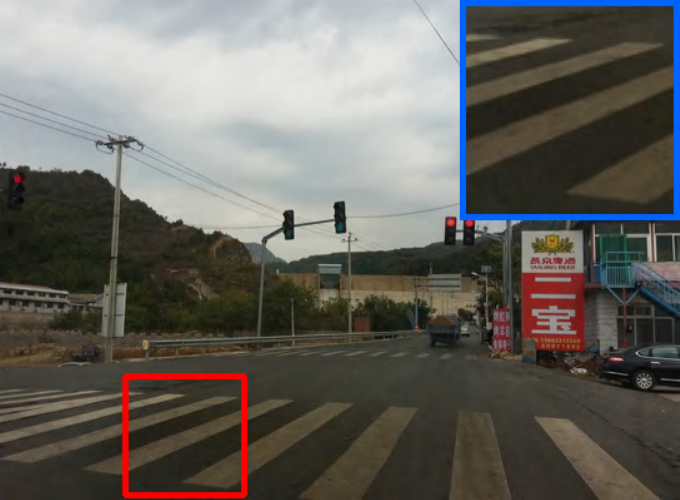}}
    \vspace{2pt}
    \centerline{\includegraphics[width=\textwidth]{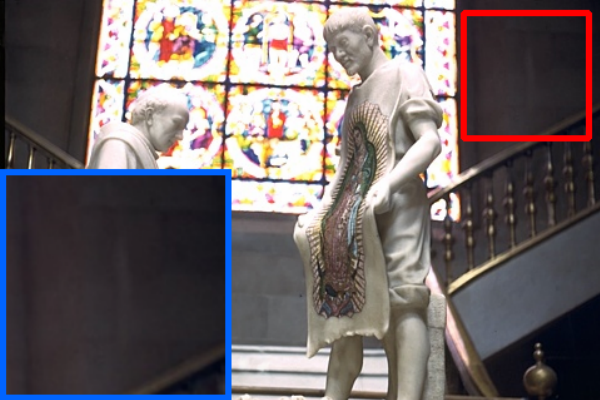}}
    \vspace{2pt}
    \centerline{\includegraphics[width=\textwidth]{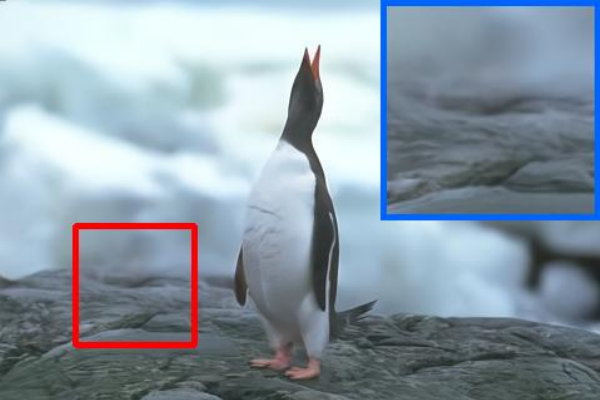}}
    \centerline{\small StarIR}
\end{minipage}
\begin{minipage}[t]{0.135\linewidth}
    \centering
    \centerline{\includegraphics[width=\textwidth]{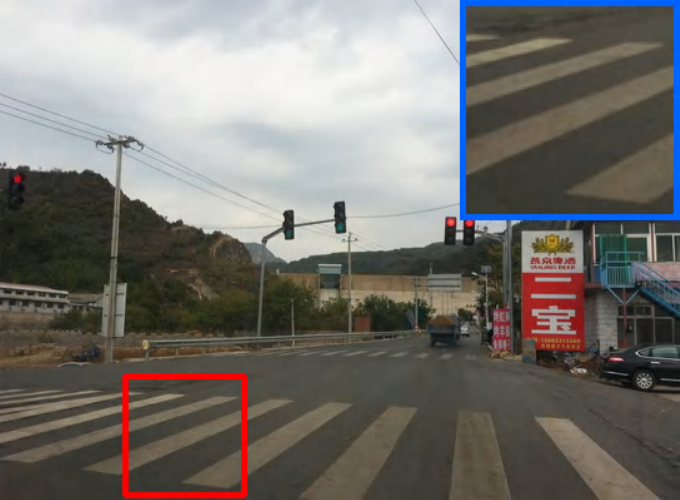}}
    \vspace{2pt}
    \centerline{\includegraphics[width=\textwidth]{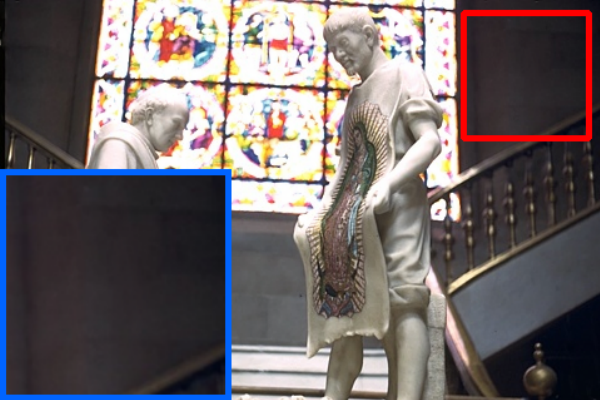}}
    \vspace{2pt}
    \centerline{\includegraphics[width=\textwidth]{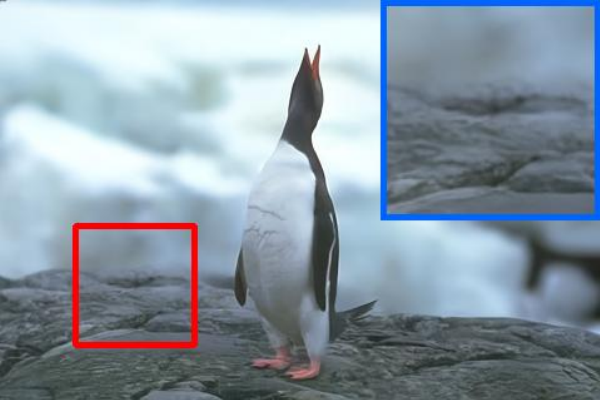}}
    \centerline{\small Ours}
\end{minipage}
\begin{minipage}[t]{0.135\linewidth}
    \centering
    \centerline{\includegraphics[width=\textwidth]{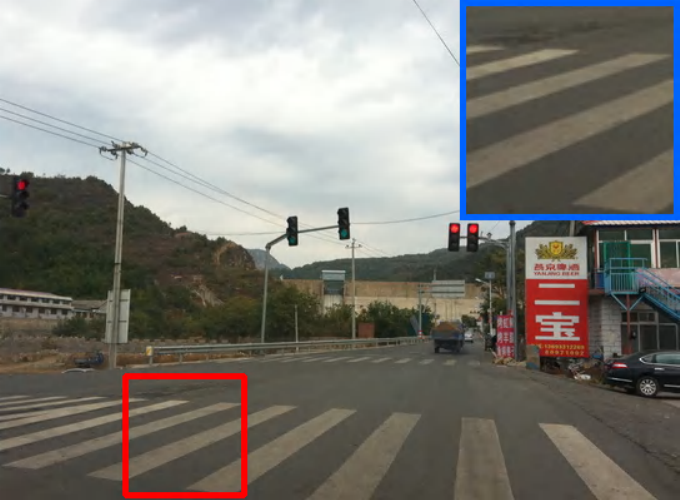}}
    \vspace{2pt}
    \centerline{\includegraphics[width=\textwidth]{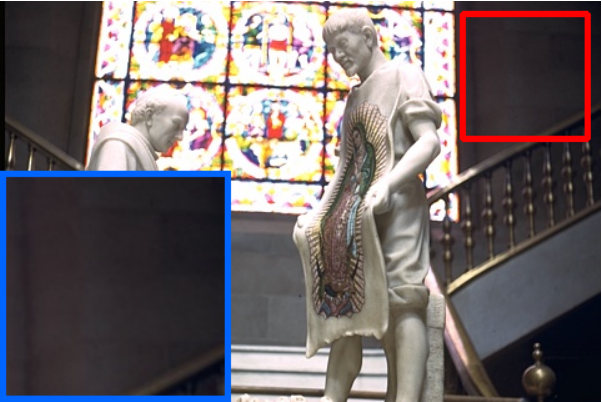}}
    \vspace{2pt}
    \centerline{\includegraphics[width=\textwidth]{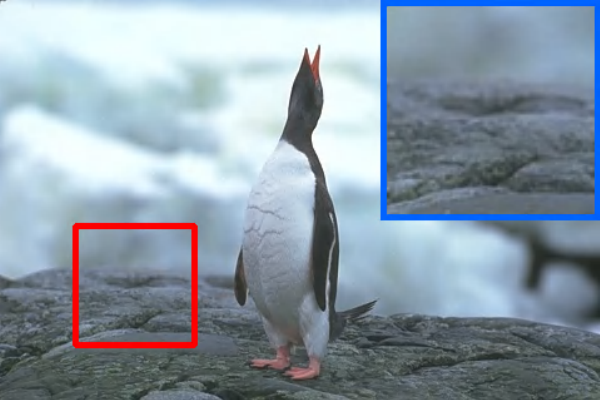}}
    \centerline{\small Ground Truth}
\end{minipage}
}
\caption{\footnotesize Visual comparisons of DPC-Net with state-of-the-art all-in-one methods on the three degradation tasks.}
\label{Fig: Three Degradations}
\vspace{-2mm}
\end{figure*}

\subsection{All-in-One Image Restoration Results}
\textbf{Three Degradation Tasks.}
We evaluate DPC-Net on three image degradation tasks, including denoising, dehazing, and deraining. As shown in Tab.~\ref{Tab: Three Degradations}, our method achieves the best average performance and demonstrates especially significant improvements on the dehazing task. As shown in Fig.~\ref{Fig: Three Degradations}, our restoration results demonstrate superior performance in texture preservation, effective removal of haze and rain streaks, and detail enhancement. This success is attributed to the guidance of the vision-language model, which enables a deep understanding of how degradation distorts image content, and to the DPCR framework’s integration of dual prior information for synergistic image restoration. 

\textbf{Five Degradation Tasks.}
Building upon the original three degradation tasks, we extend DPC-Net to encompass five degradation types. Specifically, for the newly introduced deblurring and low-light enhancement tasks, we incorporate the GoPro and LOL datasets during training, respectively. As shown in Tab.~\ref{Tab: Five Degradations}, DPC-Net achieves state-of-the-art average performance, exhibiting particularly significant superiority in the image dehazing task. Although DPC-Net has a higher parameter count than StarIR \cite{cui2026starir}, HOGFormer \cite{wu2026gradient}, R2R \cite{wang2026retrieve}, and DRNet \cite{li2026drnet}, it delivers substantially leading performance, achieving 31.00 dB (PSNR) and 0.923 (SSIM).

\textbf{Single Degradation Task.}
As shown in Tab.~\ref{Tab: Single Task}, we evaluate DPC-Net on single image restoration tasks. For image dehazing, our method surpasses the previous state-of-the-art R2R by 0.65 dB in terms of PSNR. On the image deraining task, DPC-Net achieves the best PSNR and SSIM among all competing methods. Furthermore, DPC-Net consistently delivers the best quantitative performance on the image denoising task, thereby demonstrating its strong capability across diverse restoration scenarios.

\subsection{Ablation Study}
\begin{figure*}[!t]
\centering
\scalebox{1.0}{
\begin{minipage}[t]{0.135\linewidth}
    \centering
    \centerline{\includegraphics[width=\textwidth]{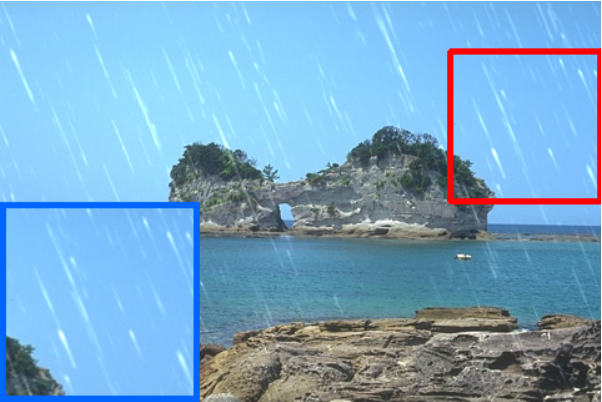}}
    \centerline{\small Input}
\end{minipage}
\begin{minipage}[t]{0.135\linewidth}
    \centering
    \centerline{\includegraphics[width=\textwidth]{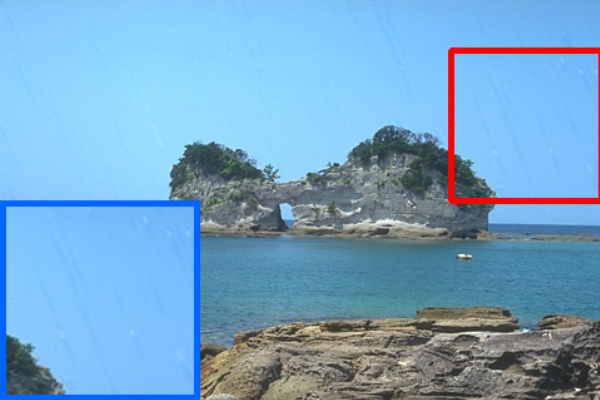}}
    \centerline{\small Variant (a)}
\end{minipage}
\begin{minipage}[t]{0.135\linewidth}
    \centering
    \centerline{\includegraphics[width=\textwidth]{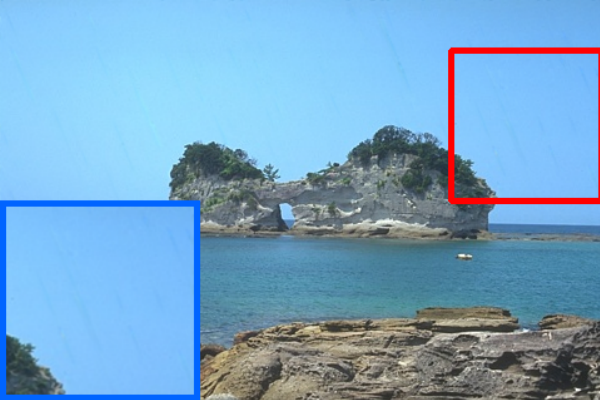}}
    \centerline{\small Variant (b)}
\end{minipage}
\begin{minipage}[t]{0.135\linewidth}
    \centering
    \centerline{\includegraphics[width=\textwidth]{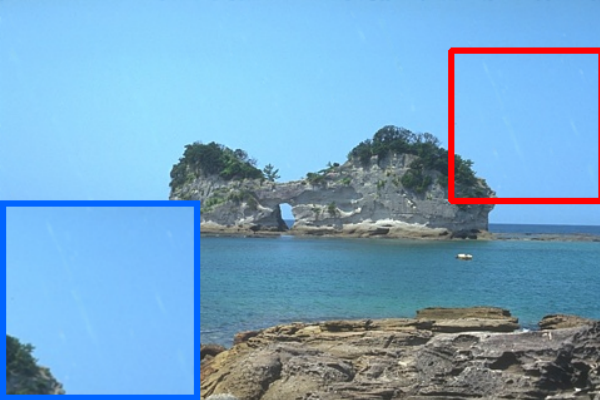}}
    \centerline{\small Variant (c)}
\end{minipage}
\begin{minipage}[t]{0.135\linewidth}
    \centering
    \centerline{\includegraphics[width=\textwidth]{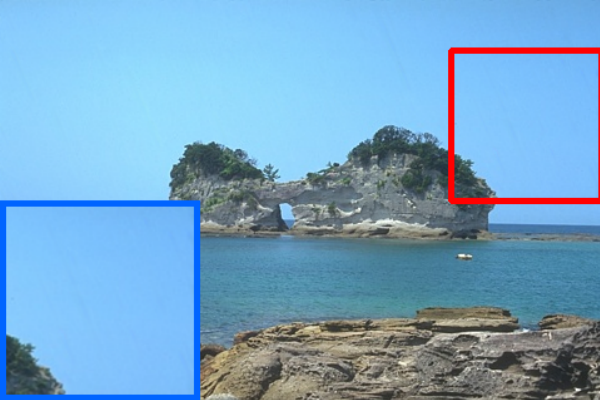}}
    \centerline{\small Variant (d)}
\end{minipage}
\begin{minipage}[t]{0.135\linewidth}
    \centering
    \centerline{\includegraphics[width=\textwidth]{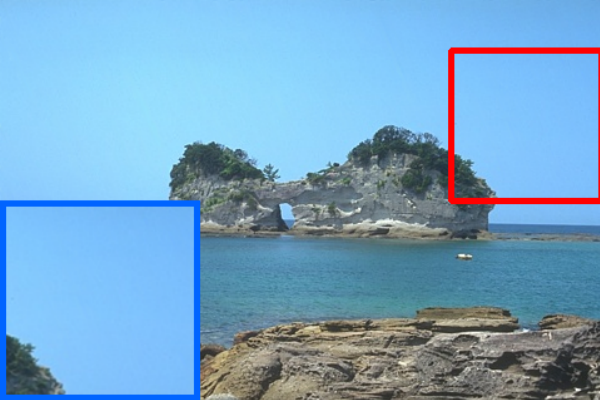}}
    \centerline{\small Ours}
\end{minipage}
\begin{minipage}[t]{0.135\linewidth}
    \centering
    \centerline{\includegraphics[width=\textwidth]{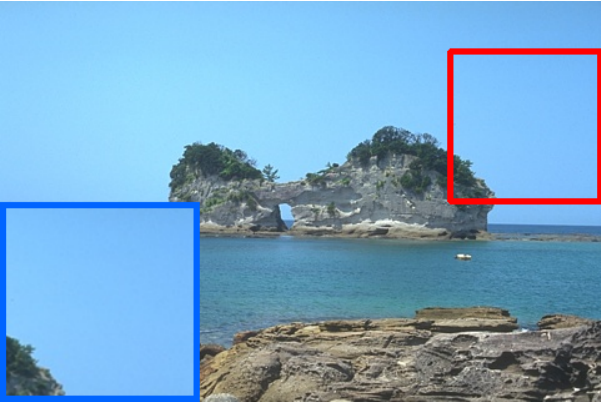}}
    \centerline{\small Ground Truth}
\end{minipage}
}
\caption{\footnotesize Visual comparisons of DPC-Net with its variants to demonstrate the progressive improvement brought by each component.}
\label{Fig: Ablation Components}
\end{figure*}

\begin{figure}[!t]
\centering
\scalebox{0.9}{
\begin{minipage}[t]{0.47\linewidth}
    \centering
    \centerline{\includegraphics[width=\textwidth]{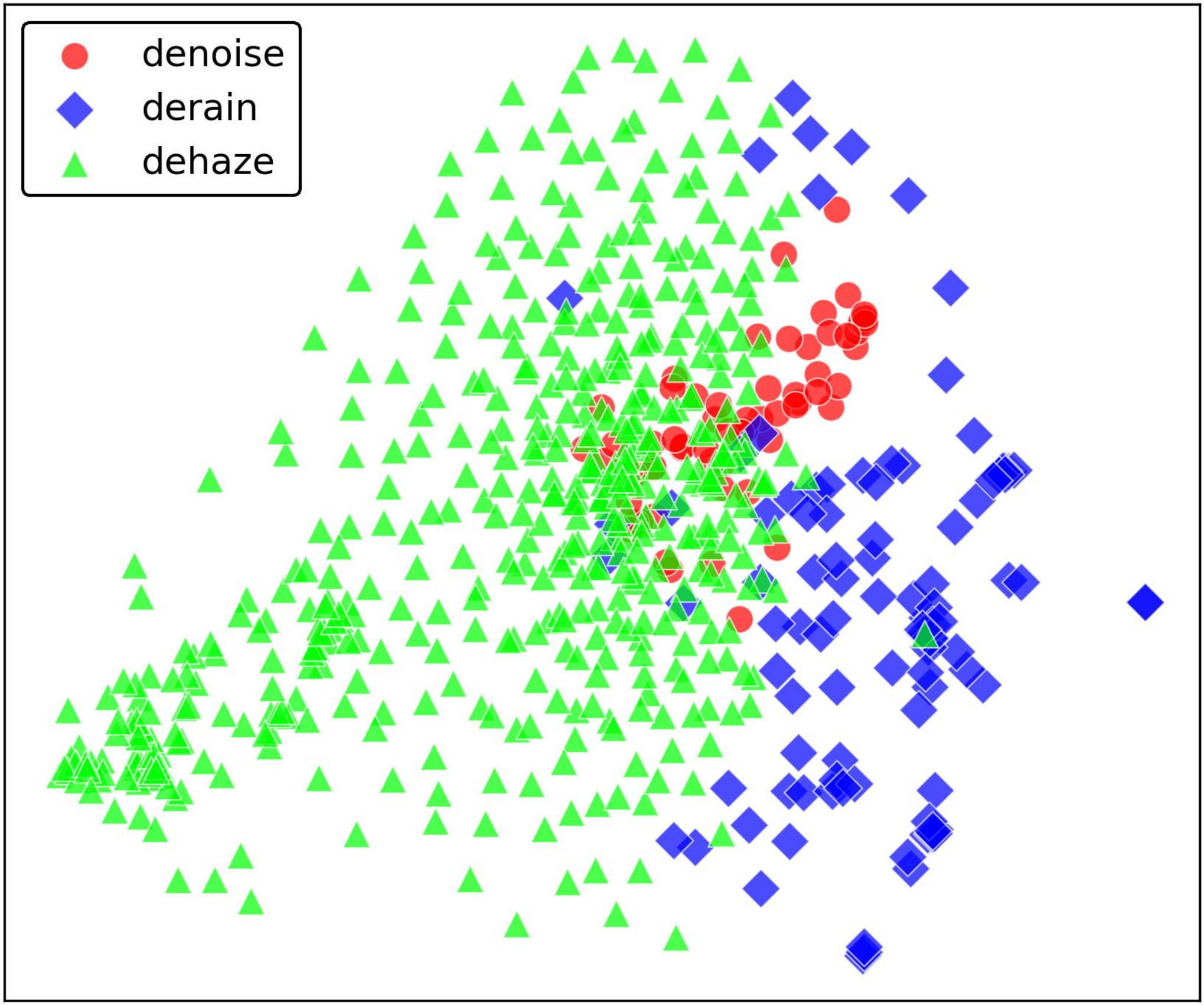}}
    \centerline{\small Variant (a)}
\end{minipage}
\begin{minipage}[t]{0.47\linewidth}
    \centering
    \centerline{\includegraphics[width=\textwidth]{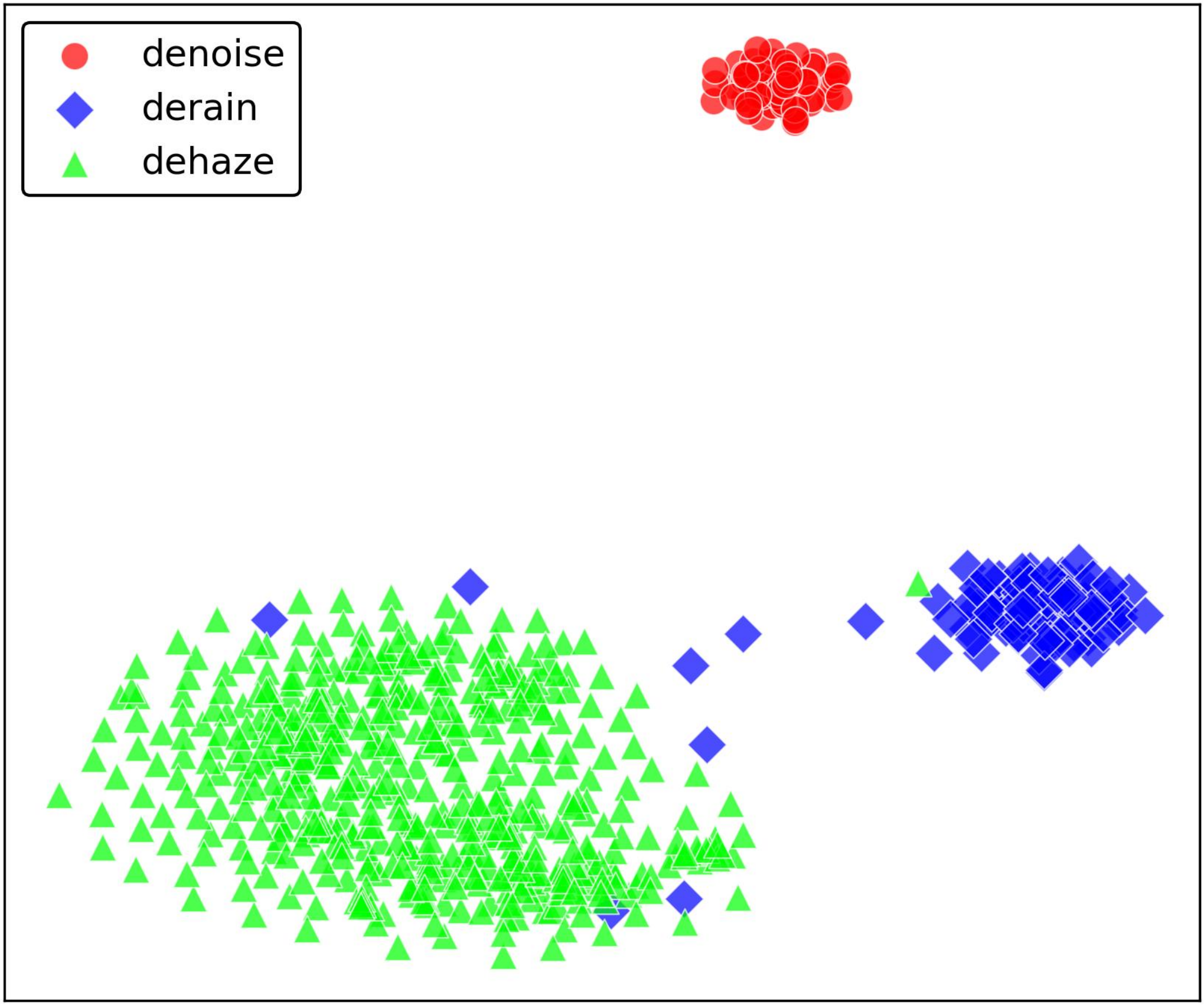}}
    \centerline{\small Ours}
\end{minipage}
}
\caption{\footnotesize t-SNE~\cite{van2008visualizing} visualization of $\mathbf{E}^n$ under the three-task setting. Variant (a) corresponds to Tab.~\ref{Tab: Ablation Components}.}
\label{Fig: t-SNE}
\vspace{-3mm}
\end{figure}

\textbf{Effects of Key Components.}
As shown in Tab.~\ref{Tab: Ablation Components}, we conduct ablation studies to quantify the contribution of each component. Specifically, Variant (a) removes DAN and DEM. Variant (b) excludes DEM while directly feeding DAN features into DPCR. Variant (c) omits the external knowledge base and DPCR, while Variant (d) removes DPCR and directly injects low-level visual priors into the decoder. Comparing Variants (a) and (b) shows that incorporating DAN significantly improves performance, demonstrating its effectiveness in learning degradation-semantic coupled features. Adding DEM further enhances performance by introducing fine-grained spatial information through hierarchical feature fusion. The t-SNE visualizations in Fig.~\ref{Fig: t-SNE} further confirm that DAN and DEM enable the encoder to better distinguish different degradation types. Furthermore, the comparison between Variants (c) and (d) indicates that the external knowledge base provides beneficial low-level visual priors for restoration. Building upon this, DPCR effectively fuses low-level visual priors with degradation semantic priors, achieving the best overall performance. As shown in Fig.~\ref{Fig: Ablation Components}, the visual results further highlight the progressive improvement brought by each component.

\begin{table*}[!htb]
\centering
\small
\setlength{\tabcolsep}{3.0pt}
\renewcommand\arraystretch{1.1}{
\resizebox{\linewidth}{!}{
\begin{tabular}{lc |lc |lc}
\toprule[1pt]
\multirow{2}{*}{Method} 
& {\textit{Dehazing}}
& \multirow{2}{*}{Method}
& {\textit{Deraining}}
& \multirow{2}{*}{Method}
& {\textit{Denoising}} \\
\cmidrule(lr){2-2}\cmidrule(lr){4-4}\cmidrule(lr){6-6}
&
{SOTS}
&
&
{Rain100L} 
&
&
{BSD68\textsubscript{$\sigma$=25}} \\
\midrule
MSCNN \cite{ren2016single} & 22.06/.908
& UMR \cite{yasarla2019uncertainty} & 32.39/.921
& CBM3D \cite{dabov2007color} & 30.69/.868
\\
AODNet \cite{li2017aod}  & 20.29/.877
& SIRR \cite{wei2019semi} & 32.37/.926
& DnCNN \cite{zhang2017beyond} & 31.23/.883
\\
EPDN \cite{qu2019enhanced} & 22.57/.863
& MSPFN \cite{jiang2020multi} & 33.50/.948
& IRCNN \cite{zhang2017learning} & 31.18/.882
\\
FDGAN \cite{dong2020fd} & 23.15/.921
& LPNet \cite{gao2019dynamic} & 23.15/.921
& BRDNet \cite{tian2020image} & 31.43/.885
\\
\midrule
AirNet \cite{li2022all}  & 23.18/.900
& AirNet \cite{li2022all} & 34.90/.977
& AirNet \cite{li2022all} & 31.48/.893
\\
PromptIR \cite{potlapalli2023promptir} & 31.31/.973
& PromptIR \cite{potlapalli2023promptir} & 37.04/.979
& PromptIR \cite{potlapalli2023promptir} & \textcolor{blue}{31.71/.897}
\\
R2R \cite{wang2026retrieve} & \textcolor{blue}{31.50/.978}
& R2R \cite{wang2026retrieve} & \textcolor{blue}{37.45/.980}
& R2R \cite{wang2026retrieve} & 31.64/.897
\\
\midrule
Ours & \textcolor{red}{32.15/.982}
& Ours & \textcolor{red}{37.65/.981}
& Ours & \textcolor{red}{31.73/.900}
\\
\bottomrule[1pt]
\end{tabular}}}
\caption{\footnotesize Comparison to state-of-the-art all-in-one methods on the single degradation task. \textcolor{red}{Best} and \textcolor{blue}{second} best performances are highlighted. PSNR (dB, $\uparrow$) and SSIM ($\uparrow$) metrics are reported on the full RGB images.}
\label{Tab: Single Task}
\end{table*}

\begin{table}[t] 
\small
\centering
\setlength\tabcolsep{8.5pt}
\renewcommand\arraystretch{1.1}{
\begin{tabular}{ccccc|cc}
\toprule[1pt]
Index & (1) & (2) & (3) & (4) & PSNR$\uparrow$ & SSIM$\uparrow$ \\
\midrule
(a) & \ding{55} & \ding{55} & \ding{51} & \ding{51} & 32.65 & 0.920  \\
(b) & \ding{51} & \ding{55} & \ding{51} & \ding{51} & 32.86 & 0.920  \\
(c) & \ding{51} & \ding{51} & \ding{55} & \ding{55} & 32.66 & 0.919  \\
(d) & \ding{51} & \ding{51} & \ding{51} & \ding{55} & 32.79 & 0.920  \\
Ours & \ding{51} & \ding{51} & \ding{51} & \ding{51} & \textcolor{red}{33.01} & \textcolor{red}{0.922} \\
\bottomrule[1pt]
\end{tabular}}
\caption{\footnotesize Effectiveness of key components under the three degradation tasks. (1), (2), (3), and (4) denote DAN, DEM, knowledge base, and DPCR, respectively.}
\label{Tab: Ablation Components} 
\end{table}

\begin{table}[t] 
\small
\centering
\setlength\tabcolsep{13.5pt}
\renewcommand\arraystretch{1.1}{
\begin{tabular}{cc|cc}
\toprule[1pt]
Index & Method & PSNR$\uparrow$ & SSIM$\uparrow$ \\
\midrule
(a) & Prompt & 32.46 & 0.918  \\
(b) & DA-CLIP & 32.64 & 0.919  \\
(c) & LLaVA (Ours) & \textcolor{red}{33.01} & \textcolor{red}{0.922} \\
\bottomrule[1pt]
\end{tabular}}
\caption{\footnotesize Effectiveness of LLaVA-based guidance strategy under the three degradation tasks.}
\label{Tab: VLM Guidance}
\end{table}


\begin{table}[t]
\centering
\fontsize{9}{11}\selectfont
\setlength\tabcolsep{11.5pt}
\renewcommand\arraystretch{1.0}
\begin{tabular}{c|cc|cc}
\toprule[1pt]
\multirow{2}{*}{$m$}
& \multicolumn{2}{c|}{w/o QGM}
& \multicolumn{2}{c}{w QGM} \\
\cline{2-5}
& PSNR$\uparrow$ & SSIM$\uparrow$
& PSNR$\uparrow$ & SSIM$\uparrow$ \\
\midrule
128 & 32.62 & 0.918 & 32.73 & 0.920 \\
256 & 32.75 & 0.920 & \textcolor{red}{33.01} & \textcolor{red}{0.922} \\
512 & 32.64 & 0.920 & 32.81 & 0.920 \\
\bottomrule[1pt]
\end{tabular}
\caption{\footnotesize Effectiveness of the QGM and comparison of knowledge base capacities across three degradation tasks.}
\label{Tab: Knowledge Base}
\vspace{-2mm}
\end{table}

\textbf{Effects of LLaVA-based Guidance Strategy.}
As shown in Tab.~\ref{Tab: VLM Guidance}, we conduct an ablation study to validate the efficacy of our LLaVA-based guidance strategy. The prompt-based approach yields the lowest performance, primarily due to its shallow level of semantic guidance, which hinders the establishment of deep correlations between degradation features and visual content. In contrast, the DA-CLIP-based strategy incorporates structured semantic constraints and achieves moderate improvement. However, this approach focuses solely on degradation type discrimination while failing to fully exploit the deep semantic priors embedded within image content. Our method attains optimal results by leveraging LLaVA's robust cross-modal understanding capability. Specifically, textual descriptions generated by LLaVA not only identify degradations but also guide the model in comprehending how such degradations visually distort specific image content. Consequently, these detailed semantic contexts significantly enhance restoration quality.

\textbf{Effects of QGM and Knowledge Base Capacity.}
As shown in Tab.~\ref{Tab: Knowledge Base}, we evaluate the effectiveness of QGM and investigate the impact of knowledge base capacity. Experimental results demonstrate that incorporating the QGM module significantly enhances model performance by leveraging degraded semantic coupling features to effectively filter out noise interference in decoder-level features, thereby generating clean queries. Regarding knowledge base capacity, we compare settings of $128$, $256$, and $512$ feature vectors. The model achieves optimal performance with $256$ vectors, indicating this scale strikes an optimal balance between representational capacity and computational complexity.

\section{Conclusion}
\label{conclusion}
This paper presents DPC-Net, a Dual-Prior Collaborative Network for all-in-one image restoration. 
Specifically, the Degradation-Aware Network extracts degradation-semantic coupled features under the supervision of a Vision-Language Model, which constrains their feature distribution. The Degradation-Semantic Modulation Module then converts this semantic guidance into degradation-semantic coupling and propagates the coupled representations to the decoder. During decoding, the Dual-Prior Collaborative Reconstruction Module fuses degradation semantic priors with low-level visual priors from knowledge bases, enabling effective restoration while preserving texture and structural fidelity. 
Extensive experiments demonstrate that DPC-Net consistently outperforms state-of-the-art AiOIR methods.


\bibliography{aaai2027}

\end{document}